%% file: main.tex
\documentclass[letterpaper, 10 pt, conference]{ieeeconf}  
\IEEEoverridecommandlockouts                              
\usepackage{epsfig} 
\usepackage{mathptmx} 
\usepackage{times} 
\usepackage{amsmath} 
\usepackage{amssymb}  
\usepackage{graphicx}
\usepackage{capt-of}
\usepackage{cite}
\usepackage{float}
\usepackage{hyperref}
\usepackage{xcolor}
\definecolor{darkgreen}{RGB}{0,100,0}
\definecolor{darkblue}{RGB}{0,51,102}
\definecolor{niceblue}{rgb}{0.0,0.4,0.8}
\definecolor{riceblue}{HTML}{00205B}
\usepackage{xcolor}
\usepackage{xurl}
\usepackage{hyperref}

\definecolor{riceblue}{HTML}{00205B}

\title{\LARGE \bf
TMA: Sim-to-Real Forceful Humanoid Manipulation \\via \underline{T}actile-Driven \underline{M}otor \underline{A}daptation}

\IEEEaftertitletext{%
\vspace{-1em}
\begin{minipage}{\textwidth}
    \centering
    \includegraphics[width=0.95\textwidth]{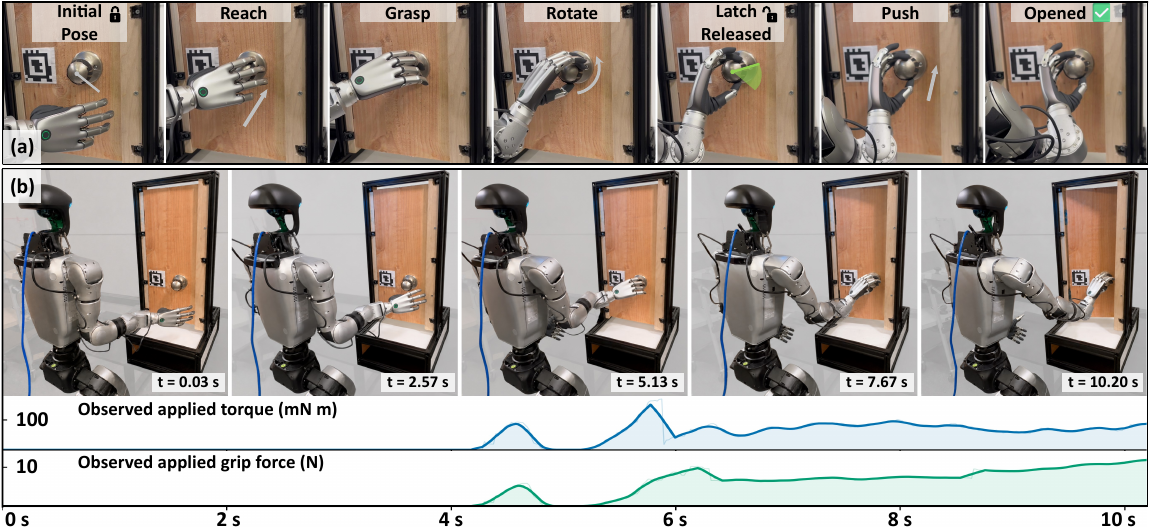}
\captionof{figure}{
\textbf{Forceful humanoid articulated manipulation for round-doorknob opening.} \textbf{(a)} The task requires tightly coupled stages: reaching, establishing a frictional multi-finger grasp, generating sufficient torque for latch release, and pushing the door open. \textbf{(b)} A representative rollout of the proposed system shows zero-shot sim-to-real execution after simulation-only training, with whole-body coordination throughout the task. Time-aligned real-world tactile readings quantify the forceful interaction through the aggregate fingertip grip force (\textcolor{darkgreen}{\textbf{green}}) and the estimated torque applied about the knob axis (\textcolor{darkblue}{\textbf{blue}}).
}
    \label{fig:teaser}
\end{minipage}
\vspace{0.8em}
}
\author{
    Yiting Chen$^{\dagger,2}$, Kelin Yu$^{1,3}$,
    Sarthak Kaingade$^{1}$, Bryan Trinh$^{1}$, \\
    Kaiyu Hang$^{2}$, Roberto Martín-Martín$^{1,4}$,
    Nima Fazeli$^{1,5}$, X. Alice Wu$^{1}$ \\[0.5em]
    {\normalsize $^{1}$Amazon \quad $^{2}$Rice University \quad
    $^{3}$University of Maryland}  \\
    {\normalsize $^{4}$The University of Texas at Austin \quad
    $^{5}$University of Michigan}  \\
    {\normalsize Project page: \href{https://humanoid-tma.github.io/}{\color{riceblue}{\textbf{humanoid-tma.github.io}}}}
    \thanks{$^{\dagger}$Work done during an internship at Amazon}
}

\begin{document}

\maketitle
\thispagestyle{empty}
\pagestyle{empty}

\begin{abstract}

Recent humanoid loco-manipulation in articulated environments has largely focused on geometry-dominant interactions, where success is primarily determined by geometric alignment and motion execution. In contrast, forceful multi-finger manipulation requires the robot to adapt to real-world uncertainties in contact and interaction dynamics. We present Tactile-Driven Motor Adaptation (TMA) and an integrated whole-body system for zero-shot sim-to-real forceful humanoid manipulation, and study round-doorknob opening as a challenging benchmark requiring stable multi-finger frictional grasping, sufficient torque generation without slip, force-regulated pushing, and whole-body stability. TMA structures policy learning around interactions observable through tactile feedback, uses cone geometry as an inductive bias for multi-finger contact representation, and adapts motor behavior online from tactile history. An upper-body manipulation policy is trained alongside a pretrained general locomotion controller to coordinate dexterous interaction and whole-body stabilization. Trained entirely in simulation, TMA transfers zero-shot to a Unitree G1 humanoid with BrainCo Revo2 multi-fingered hands and substantially outperforms standard Rapid Motor Adaptation (RMA) and reactive tactile policy baselines across four commercial round doorknobs in the real world.  

\end{abstract}


\input{text/01-introduction}


\input{text/02-related-work}


\input{text/03-method}


\input{text/04-experiments}


\input{text/05-results}






\bibliographystyle{IEEEtran}
\bibliography{references}

\end{document}

%% file: text/01-introduction.tex
\section{INTRODUCTION}

Dexterous manipulation is essential for humanoids to physically operate in human environments~\cite{billard2019trends, chen2026manipulationnet}. Recent data-driven approaches have rapidly advanced humanoid loco-manipulation, enabling increasingly complex interactions with objects and articulated environments~\cite{gu2026humanoid}. However, these systems are predominantly designed in what we term a \textit{geometry-dominant} manner: success is determined mainly by where contact is established and how the robot moves with respect to object geometry and kinematic constraints, without explicitly regulating the \textit{evolving physical state of contact}~\cite{hogan2022contact}. Consider door opening by a humanoid equipped with a multi-fingered hand. A lever handle can often be actuated by establishing appropriate contact and exploiting its favorable geometry and moment arm~\cite{xue2026opening, doorbot}. A round doorknob, however, exposes an \textit{interaction-dominant} regime, where success depends critically on how fingers establish contact, distribute normal forces, generate sufficient frictional torque, and avoid slip~\cite{holladay}. Such tasks require not only accurate motion, but continuous regulation and adaptation of the increasing resisting torque with the evolving contact state. Enabling humanoids to perform this class of dexterous manipulation remains largely unexplored.

In this work, we investigate interaction-dominant manipulation through round-doorknob opening, as shown in Fig.~\ref{fig:teaser}. In this task, action outcomes depend critically on physical properties not directly observable from vision or proprioception, including contact friction and internal mechanism dynamics such as spring stiffness and latch resistance. Before their physical consequences unfold, a stable grasp and an imminent slip may induce nearly indistinguishable visual and proprioceptive observations, yet lead to drastically different outcomes under the same motion command. Therefore, this observability gap makes tactile sensing particularly valuable, as it directly measures the local contact response induced by the robot's actions~\cite{li2020review}. This calls for a tactile-driven policy that treats contact not only as an actuation interface, but also as an information source for regulating interaction.

However, tactile feedback alone does not resolve interaction uncertainty. While a reactive tactile policy can respond to instantaneous contact, many task-relevant physical properties are only revealed through how contact evolves under the robot's actions. The same instantaneous contact force, for example, may correspond to different frictional margins or mechanism responses depending on the preceding interaction. Effective interaction-dominant manipulation therefore requires accumulating evidence through physical interaction, inferring the underlying interaction conditions, and adapting motor behavior accordingly.

To this end, we introduce Tactile-Driven Motor Adaptation (TMA), a framework for interaction-dominant humanoid manipulation under real-world contact uncertainty. While retaining a history-conditioned adaptation structure, TMA departs from the Rapid Motor Adaptation (RMA)~\cite{kumar2021rma} paradigm by shifting adaptation from observations accumulated during execution to interaction-driven observations deliberately acquired through active contact. Rather than only inferring environmental properties from the consequences encountered during policy execution, TMA exploits physical interaction to elicit tactile responses that reveal task-relevant interaction conditions. Contact thus serves a dual role: it is both the interface through which the robot acts on the environment and the sensing mechanism through which otherwise unobserved physical properties become inferable. We instantiate this principle by 1) structuring policy learning around tactile-observable interactions, 2) parameterizing tactile signals through cone geometry to ground policy learning in contact mechanics, and 3) adapting motor behavior online from recent tactile interaction history. The policy and adaptation encoder are trained entirely in simulation using reinforcement learning and supervised latent regression, then transferred zero-shot to the real-world humanoid system. At the system level, we integrate TMA with a frozen general locomotion controller, enabling forceful round-doorknob manipulation while maintaining whole-body stability.

We evaluate TMA on zero-shot sim-to-real round-doorknob opening using a Unitree G1\footnote{https://www.unitree.com/g1} humanoid equipped with BrainCo Revo2\footnote{https://www.brainco-hz.com/docs/revolimb-hand/en/} dexterous hands. Across four commercially available doorknob models, TMA substantially outperforms Rapid Motor Adaptation and reactive tactile policy baselines. To the best of our knowledge, this is the first demonstration of this task by a humanoid using a multi-fingered dexterous hand. Our contributions are threefold:
\begin{itemize}
\item We study round-doorknob opening as a challenging benchmark for interaction-dominant dexterous humanoid manipulation, requiring coordinated multi-finger force interaction and whole-body stability.
\item We propose Tactile-Driven Motor Adaptation (TMA), an adaptive policy framework that exploits active tactile interaction history for interaction regulation and enables zero-shot sim-to-real transfer.
\item We develop an integrated whole-body humanoid system that combines dexterous tactile manipulation with a general locomotion controller and demonstrates robust real-world doorknob opening.
\end{itemize}

%% file: text/02-related-work.tex
\section{RELATED WORK}
\noindent\textbf{Learning-based Humanoid Manipulation.} Learning-based humanoid manipulation has advanced rapidly, enabling increasingly diverse and complex whole-body interaction skills~\cite{gu2026humanoid}. Humanoid teleoperation maps operator commands to robot motion~\cite{he2024omniho,ze2025twist,ben2025homie,lu2025mobile}, allowing humans to guide task execution and collect demonstrations for policy learning~\cite{he2024omniho,fu2024humanplus}. Imitation learning uses teleoperated or video demonstrations to acquire autonomous manipulation policies~\cite{fu2024humanplus,li2024okami}, and can be combined with RL-trained whole-body controllers to support coordinated execution~\cite{he2024omniho,fu2024humanplus}. In parallel, large-scale simulation enables reinforcement learning of task policies through trial-and-error, with sim-to-real transfer supporting object transport~\cite{dao2024box, zhao2025resmimic}, dexterous grasping~\cite{lin2025simtoreal}, long-horizon loco-manipulation~\cite{he2026viral}, and articulated-environment interaction~\cite{xue2026opening}. Force-adaptive approaches further coordinate upper-body tracking and lower-body stability to sustain manipulation under substantial end-effector loads~\cite{zhang2025falcon}.

However, these advances still model physical interaction predominantly at the level of motion, end-effector forces, or whole-body response, leaving finger-level forceful interaction underexplored. Our research targets forceful multi-fingered humanoid manipulation, where task success depends on the contact states and forces distributed across individual fingers and their evolution during interaction.

\vspace{3pt}
\noindent\textbf{Sim-to-Real Transfer of Tactile Policies.} Tactile sensing has increasingly enabled sim-to-real manipulation by exposing contact information not directly observable from vision and proprioception alone~\cite{qi2023general,yang2024anyrotate}. Prior work has transferred tactile policies from simulation to real robots for door manipulation~\cite{ding2021sim,lee2024dextouch}, tactile servoing and bimanual manipulation~\cite{church2022tactile,lin2023bi}, dexterous in-hand manipulation~\cite{yang2024anyrotate,yuan2024robot,yang2023tacgnn}, and multi-finger manipulation with dense force feedback~\cite{zhao2026closing}. These approaches bridge the tactile sim-to-real gap through sensor modeling~\cite{dang2026hydroshear}, observation translation~\cite{church2022tactile,freud2025simshear}, latent alignment~\cite{yang2026tactile}, and robust policy learning with domain randomization~\cite{ding2021sim,yang2024anyrotate}. Complementarily, motor adaptation enables policies to infer latent representations of unobserved interaction conditions from recent history and adjust behavior online~\cite{kumar2021rma,qi2023hora}. More closely related to our setting, tactile feedback has also been incorporated into history-conditioned adaptation and robust policy learning for generalizable in-hand manipulation, complementing proprioception~\cite{yin2025learning,yang2024anyrotate} and vision~\cite{qi2023general}.

However, tactile policies based on discretized contact cues~\cite{qi2023general,yin2025learning} or scalar force magnitudes~\cite{yang2024anyrotate} capture only a partial view of the underlying force state. Our research preserves a force-aware contact state encoding both magnitude and direction, and adapts motor behavior online from its temporal evolution across multi-fingered interactions.

%% file: text/03-method.tex
\section{PRELIMINARIES \& PROBLEM STATEMENT}
\label{sec:problem}

We consider round-doorknob opening by a floating-base humanoid equipped with multi-fingered dexterous hands. Starting from a standing pose in front of a closed door, the robot must reach the knob, establish a frictional grasp, rotate the knob sufficiently to release the latch, and move the door panel to an open configuration without losing balance. The task therefore couples arm motion, multi-finger contact, doorknob-mechanism actuation, and whole-body stability.

\subsection{Preliminaries}
\label{sec:preliminaries}

\begin{figure}[H]
    \centering

    \includegraphics[width=0.8\columnwidth]{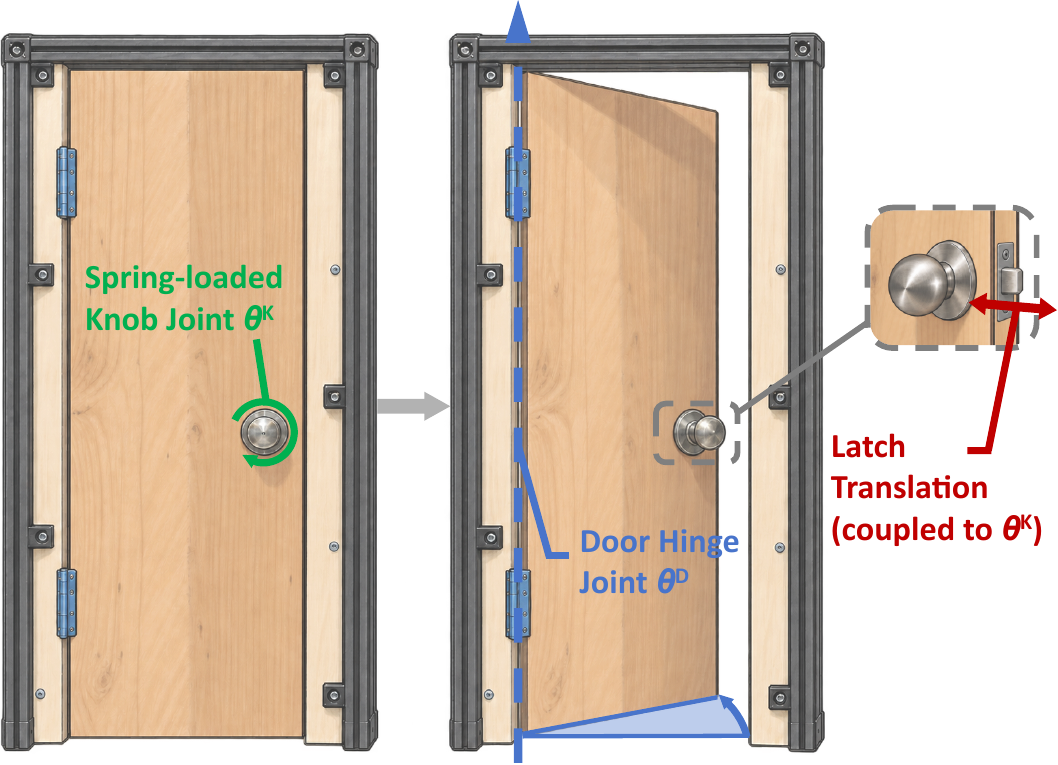}
\caption{
\textbf{Articulated door mechanism.}
The door model comprises a spring-loaded knob joint $\theta^K$, a revolute hinge joint $\theta^D$, and a translating latch coupled to knob rotation. Knob rotation retracts the latch: once $\theta^K$ reaches the release threshold $\theta_{\mathrm{rel}}$, the door panel can rotate about the hinge.}
    \label{fig:prelim}
        \vspace{-1em}

\end{figure}

\subsubsection{Coordinate Frames and Task Geometry}
Let $\{W\}$ denote an inertial world frame and $\{P\}$ a pelvis frame attached to the robot's pelvis. The knob geometry is described by its center $\mathbf p_{\{PK\}}\in\mathbb{R}^3$, unit axle direction $\mathbf d_{\{PK\}}\in\mathbb{S}^2$, and rim radius $r_K>0$. The subscript $\{PK\}$ denotes knob quantities expressed in the pelvis frame, and $\mathbb{S}^2$ denotes the unit sphere in $\mathbb{R}^3$.

As shown in Fig.~\ref{fig:prelim}, the door mechanism has a hinge angle $\theta^D$ and a knob angle $\theta^K$, both measured from their closed positions. Positive hinge rotation corresponds to door opening, while positive knob rotation corresponds to latch retraction. The knob center and axle move rigidly with the panel, while the knob rotates about its axle to retract the latch. The knob joint is spring-loaded, with a spring torque that opposes rotation in the release direction:
\begin{equation}
    \tau^K=-\bigl(k_K\theta^K+\tau_K^0\bigr),
    \qquad \theta^K\geq0 
    \label{eq:knob_spring}
\end{equation}
Here, $k_K>0$ is the spring stiffness, which determines the increase in opposing torque per unit of knob rotation. The preload $\tau_K^0\geq0$ is the magnitude of the closing torque already present at $\theta^K=0$. The negative sign indicates that the spring acts toward the closed position. The latch releases when $\theta^K\geq\theta_{\mathrm{rel}}$, where $\theta_{\mathrm{rel}}>0$ is the release threshold.

\subsubsection{Robot Configuration and Control}
Let $\mathbf q\in Q\subset\mathbb{R}^{n_q}$ denote the measured actuated joint positions, where $Q$ is bounded by joint limits. The vector $\mathbf q$ is partitioned into manipulation joints $\mathbf q^m\in\mathbb{R}^{n_a}$, locomotion joints $\mathbf q^l\in\mathbb{R}^{n_l}$, and held joints $\mathbf q^h\in\mathbb{R}^{n_h}$, with $n_q=n_a+n_l+n_h$. The manipulation policy controls the right arm and hand joints $\mathbf q^m$, while the locomotion controller commands the leg and waist joints $\mathbf q^l$. The held joints $\mathbf q^h$ track fixed reference targets. The manipulation action $\mathbf a\in A=\mathbb{R}^{n_a}$ determines joint-position increments that are integrated into position targets.

\subsubsection{Tactile Sensing}
Let $n_f\in\mathbb{N}_{>0}$ denote the number of fingertips selected for tactile feedback. A local tactile frame $\{T_i\}$ is attached to each fingertip link $i\in\{1,\ldots,n_f\}$, with a fixed position and orientation relative to that link. Its origin is at the sensing location, its $+Z$-axis follows the outward pad normal, and its $XY$-plane is tangent to the pad surface. The rotation $R_{\{WT_i\}}\in\mathrm{SO}(3)$ describes the tactile-frame orientation in the world frame.

The sensor output is a force vector expressed in $\{T_i\}$:
\begin{equation}
    \mathbf f_i=[f_i^x,f_i^y,f_i^z]^\top\in\mathbb{R}^3
    \label{eq:raw_tactile}
\end{equation}
The force sign corresponds to the fingertip acting on the object, so $f_i^z>0$ denotes compressive loading along the pad normal. The components $f_i^x$ and $f_i^y$ are signed tangential forces along the local $X$- and $Y$-axes. The radius-$\rho$ neighborhood of the origin of $\{T_i\}$ defines fingertip $i$'s tactile footprint, with knob-contact set $C_{i,t}(\rho)$ and indicator $c_{i,t}=\mathbf{1}[|C_{i,t}(\rho)|>0]$.

\subsection{Problem Statement}

The task is formulated as a finite-horizon partially observable Markov decision process, $M=(S,A,O,T,\Omega,R,\gamma,H)$, with state, action, and observation spaces $S$, $A$, and $O$, transition dynamics $T$, and observation model $\Omega$. Control steps are indexed by $t\in\mathbb{N}_0$. The state $s_t\in S$ describes the complete robot--door interaction, including physical properties unavailable at deployment, whereas $\mathbf o_t\in O$ contains only deployable measurements, estimates, and manipulation targets. The deployed policy maps observation history to a motor command, $\mathbf a_t\sim\pi(\cdot\mid\mathbf o_{0:t})$. A policy is trained through reinforcement learning to maximize the expected return
\begin{equation}
    J(\pi)=\mathbb{E}_{\pi}\!\left[
    \sum_{t=0}^{H-1}\gamma^t R(s_t,\mathbf a_t,s_{t+1})\right]
    \label{eq:pomdp_objective}
\end{equation}
where $R$ is the task reward, $\gamma\in[0,1]$ is the discount factor, and $H$ is the episode horizon.

\section{METHOD}
\label{sec:method}
\begin{figure*}[t]
    \centering

    \includegraphics[width=\textwidth]{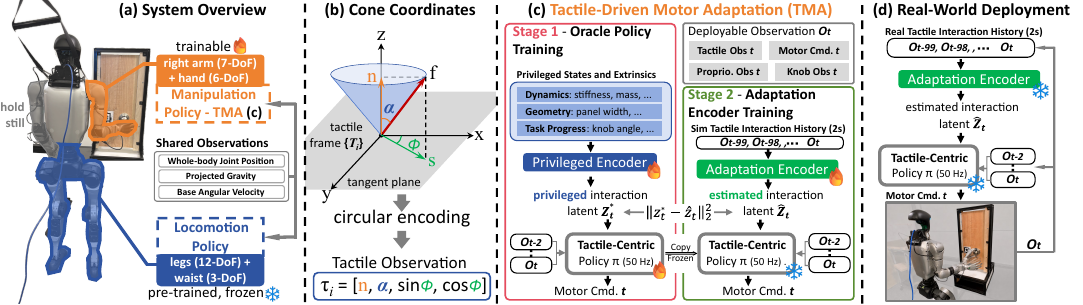}

\caption{\textbf{Overview of Tactile-Driven Motor Adaptation.}
\textbf{(a)} The upper-body manipulation policy commands the right arm and hand, while a pretrained locomotion controller commands the leg and waist joints to maintain whole-body balance under interaction-induced disturbances. The two policies share whole-body proprioceptive feedback, coupling manipulation and balance within an integrated control system.
\textbf{(b)} Each fingertip contact force is parameterized in cone coordinates by the normal load $n$, force cone angle $\alpha$, and circularly encoded shear direction $(\sin\phi,\cos\phi)$. This formulation separates the uncertainty source while preserving the contact mechanics.
\textbf{(c)} The policy learning proceeds in two stages. First, the privileged encoder and tactile-centric policy are optimized jointly end-to-end under the task objective, forming an oracle policy and yielding a task-relevant interaction latent. Both modules are then frozen, and the adaptation encoder is trained through latent regression to estimate the interaction latent from a $2\,$s history of deployable observations.
\textbf{(d)} During real-world deployment, the adaptation encoder updates the interaction latent online from tactile interaction history and conditions the frozen tactile-centric policy to generate manipulation motor commands.}

\label{fig:method}
\vspace{-1em}
\end{figure*}

We introduce Tactile-Driven Motor Adaptation (TMA), which adapts the humanoid's manipulation behavior online from the temporal response of multi-finger contact. We integrate TMA with a locomotion controller to form a whole-body humanoid control system (Sec.~\ref{sec:system}). TMA combines three components: an interaction objective for tactile-observable task progress (Sec.~\ref{sec:interaction_learning}), a contact-centric tactile representation (Sec.~\ref{sec:tactile}), and an adaptation encoder that infers a latent representation of environmental extrinsics at deployment (Sec.~\ref{sec:tma}). Together with domain randomization designed to capture real-world variability (Sec.~\ref{sec:randomization}), these components support zero-shot sim-to-real transfer.

\subsection{System Overview}
\label{sec:system}

While TMA is designed to be agnostic to the humanoid platform, we instantiate it on a Unitree G1 Edu equipped with BrainCo Revo2 hands. As shown in Fig.~\ref{fig:method}(a), a pretrained locomotion controller~\cite{ben2025homie} is kept frozen and generates position targets $\bar{\mathbf q_t}^l\in\mathbb{R}^{15}$ for the 12 leg and three waist joints under a zero velocity command to maintain balance. The manipulation policy controls the seven right-arm joints and six right-hand driver joints $\bar{\mathbf q_t}^m\in\mathbb{R}^{13}$, while the seven left-arm joints and six left-hand driver joints are held at fixed reference targets $\bar{\mathbf q_t}^h\in\mathbb{R}^{13}$.

The Revo2 hands use a joint-mimic mechanism in which each distal finger joint follows its corresponding proximal joint at a fixed ratio. Each fingertip is equipped with a capacitive tactile sensor that provides a three-dimensional contact force measurement, as defined in Eq.~\eqref{eq:raw_tactile}. 

We use mjlab~\cite{zakka2026mjlab} for rigid-body simulation as it provides direct access to low-level contact data~\cite{todorov2012mujoco}.

\subsection{Interaction-Structured Reward Formulation}
\label{sec:interaction_learning}

Fingertip contact serves both to actuate the door mechanism and to provide tactile evidence about interaction dynamics. The reward couples these roles by conditioning knob rotation and door opening rewards on an opposition grasp within the tactile sensing regions. This encourages task progress through contacts that provide tactile feedback for adaptation. With weights absorbed into the individual terms, the core task reward $r_t^{\rm task}$ is
\begin{equation}
    \begin{aligned}
    r_t^{\rm task}={}&r_t^{\rm reach}
    +w_t\bigl(r_t^{\rm knob}+r_t^{\rm hinge}\bigr)
    -r_t^{\rm pen}
    \end{aligned}
    \label{eq:interaction_reward}
\end{equation}
where $w_t$ is a grip gate. The reach term $r_t^{\rm reach}$ encourages the fingertips to approach the knob rim. The rotation terms $r_t^{\rm knob}$ and $r_t^{\rm hinge}$ reward increases in the knob and hinge angles beyond their previous episode maxima, up to their respective caps. Hinge progress is additionally gated by $\theta_t^K\geq\theta_{\mathrm{rel}}$ to ensure latch release before door opening. The penalty $r_t^{\rm pen}$ discourages large actions and abrupt changes between consecutive actions.

The grip gate is computed from simulated contact signals before sensor noise or dropout is applied. Let $c_{i,t}\in\{0,1\}$ indicate contact with the knob rim inside fingertip $i$'s sensing footprint. With $i=1$ denoting the thumb and $i=2,3,4$ the index, middle, and ring fingers, the gate is
\begin{equation}
w_t=\begin{cases}
1, & c_{1,t}=1,\;\sum_{i=2}^{4}c_{i,t}\geq2\\
\eta, & c_{1,t}=1,\;\sum_{i=2}^{4}c_{i,t}=1\\
0, & \text{otherwise}
\end{cases}.
\label{eq:grip_gate}
\end{equation}
Here, $0\leq\eta<1$ assigns reduced credit to a thumb--finger pinch, while a grasp involving the thumb and at least two opposing fingers receives full credit. Reach rewards remain ungated to encourage contact acquisition, and action penalties apply throughout the task.

\subsection{Contact-Centric Tactile Representation}
\label{sec:tactile}


\textbf{Tactile simulation.} Each fingertip sensor is approximated by aggregating simulated contact forces over its sensing footprint. Let $C_{i,t}(\rho)$ denote the hand-knob contacts involving fingertip $i$ within distance $\rho>0$ of the origin of $\{T_i\}$. The resultant force is transformed into the local tactile frame as
\begin{equation}
    \mathbf f^{\rm raw}_{i,t}=R_{\{WT_i\},t}^{\top}
    \sum_{k\in C_{i,t}(\rho)}\mathbf f_{\{W\},ik,t}
    \in\mathbb{R}^3
    \label{eq:tactile_force}
\end{equation}
where $R_{\{WT_i\},t}$ is the tactile-frame orientation defined in Sec.~\ref{sec:preliminaries}, and $\mathbf f_{\{W\},ik,t}$ is the world-frame force exerted by fingertip $i$ on the contacted object at contact $k$. This sign convention gives positive normal force under compression. The vector $\mathbf f^{\rm raw}_{i,t}$ is the simulated counterpart of the calibrated reading in Eq.~\eqref{eq:raw_tactile}. 

\textbf{Signal preprocessing.}
Each component of $\mathbf f^{\rm raw}_{i,t}$ is clipped to $\pm F_{\max}$ as $\mathbf f^{\rm cap}_{i,t}$ to suppress contact force spikes, with $F_{\max}=20\,$N chosen based on real sensor measurements. A causal low-pass filter then approximates the finite sensor bandwidth:
\begin{equation}
\begin{aligned}
\tilde{\mathbf f}_{i,t}
&=\beta\mathbf f^{\rm cap}_{i,t}
+(1-\beta)\tilde{\mathbf f}_{i,t-1}
\end{aligned}
\label{eq:tactile_filter}
\end{equation}
Here, $\beta\in(0,1]$ controls temporal smoothing. Clipping before filtering limits the contribution of force spikes to the filter state, which is reset to zero at episode boundaries. A dead zone suppresses unstable angular encodings at negligible force magnitudes. The processed force $\hat{\mathbf f}_{i,t}$ is set to zero when $\|\tilde{\mathbf f}_{i,t}\|_2\leq0.1\,$N and equals $\tilde{\mathbf f}_{i,t}$ otherwise.

\textbf{Cone-coordinate force parameterization.}
Inspired by friction-cone contact geometry~\cite{li1994mathematical}, the processed force is parameterized by its normal load, relative shear loading, and shear direction. This representation makes the normal--tangential structure of contact explicit and provides a physical interpretation for each tactile feature.

For each fingertip, the normal load $n_{i,t}$ and shear magnitude $s_{i,t}$ are computed from the processed force. The encoding $\boldsymbol\tau_{i,t}\in\mathbb{R}^4$ in Fig.~\ref{fig:method}(b) from fingertip frame $\{T_i\}$ is then given by
\begin{equation}
\begin{aligned}
n_{i,t}&=\max(\hat f_{i,t}^z,0)\\
\alpha_{i,t}&=\operatorname{atan2}(\sqrt{(\hat f_{i,t}^x)^2+(\hat f_{i,t}^y)^2},n_{i,t}) \in[0,\pi/2]\\
\phi_{i,t}&=\operatorname{atan2}(\hat f_{i,t}^y,\hat f_{i,t}^x)\\
\boldsymbol\tau_{i,t}&=\left[n_{i,t},\alpha_{i,t},
\sin\phi_{i,t},\cos\phi_{i,t}\right]^\top 
\end{aligned}
\label{eq:tactile_encoding}
\end{equation}
The circular encoding $(\sin\phi_{i,t},\cos\phi_{i,t})$ avoids angular wraparound. When the shear magnitude is smaller than $10^{-6}$N, both directional entries are explicitly set to zero because the shear direction is unreliable. With the convention $\operatorname{atan2}(0,0)=0$, a zero processed force produces an all-zero nominal encoding.

The selected fingertips are the thumb, index, middle, and ring fingers ($n_f=4$), matching the grip gate. Their encodings form the tactile observation
\begin{equation}
\boldsymbol\tau_t=
\left[\boldsymbol\tau_{1,t}^{\top},\,
\boldsymbol\tau_{2,t}^{\top},\,
\boldsymbol\tau_{3,t}^{\top},\,
\boldsymbol\tau_{4,t}^{\top}\right]^{\top}
\in\mathbb{R}^{16}
\label{eq:tactile_observation}
\end{equation}
This vector forms the tactile component of the deployable observation $\mathbf o_t$.

\subsection{Tactile-Driven Motor Adaptation}
\label{sec:tma}

\textbf{Deployable observation.} Each observation $\mathbf o_t\in O\subset\mathbb{R}^{86}$ comprises four blocks, with $\operatorname{col}(\cdot)$ denoting vertical concatenation:
\begin{equation}
    \begin{aligned}
    \mathbf o_t
        &=\operatorname{col}\bigl(\mathbf o_t^{\mathrm{prop}},\,
                 \mathbf o_t^{\mathrm{cmd}},\,
                 \mathbf o_t^{\mathrm{tact}},\,
                 \mathbf o_t^{\mathrm{knob}}\bigr)\\
    \mathbf o_t^{\mathrm{prop}}
        &=\operatorname{col}\bigl(\mathbf q_t,\,\mathbf g_{\{P\},t},\,
                 \mathbf v_{\{P\},t},\,\boldsymbol\omega_{\{P\},t}\bigr)
          \in\mathbb{R}^{50}\\
    \mathbf o_t^{\mathrm{cmd}}&=\bar{\mathbf q}_t^m\in\mathbb{R}^{13}
    \qquad
    \mathbf o_t^{\mathrm{tact}}=\boldsymbol\tau_t\in\mathbb{R}^{16}\\
    \mathbf o_t^{\mathrm{knob}}
        &=\operatorname{col}\bigl(\mathbf p_{\{PK\},t},\,
                 \mathbf d_{\{PK\},t},\,r_K\bigr)
          \in\mathbb{R}^{7}
    \end{aligned}
    \label{eq:observation}
\end{equation}
The joint vector $\mathbf q_t$ contains $n_q=41$ actuated body and hand-driver positions. Projected gravity $\mathbf g_{\{P\},t}$, base linear velocity $\mathbf v_{\{P\},t}$, and angular velocity $\boldsymbol\omega_{\{P\},t}$ are three-dimensional vectors expressed in the pelvis frame $\{P\}$. The target $\bar{\mathbf q}_t^m$ is accumulated from actions through $t-1$. 

\textbf{Privileged encoder and oracle policy learning.} As shown in Fig.~\ref{fig:method}(c), the privileged input is partitioned as
\begin{equation}
    \mathbf e_t
    =\operatorname{col}\bigl(\mathbf e_t^{\mathrm{dyn}},\,
            \mathbf e_t^{\mathrm{geo}},\,
            \mathbf e_t^{\mathrm{prog}}\bigr)
    \in E\subset\mathbb{R}^{14}
\label{eq:privileged_observation}
\end{equation}
where $\mathbf e_t^{\mathrm{dyn}}\in\mathbb{R}^{8}$ contains knob stiffness, preload, sliding and torsional friction, panel mass, and mean hand-driver proportional gain, derivative gain, and effort limit. The geometry block $\mathbf e_t^{\mathrm{geo}}\in\mathbb{R}^{2}$ contains a knob-shape indicator and the knob-to-hinge distance. The mechanism-state block $\mathbf e_t^{\mathrm{prog}}\in\mathbb{R}^{4}$ contains hinge and knob angles and velocities. The encoder $f_\varphi^{\mathrm{priv}}$ compresses these properties and states into a task-relevant interaction latent. With stacked history $\mathbf x_t=\mathbf o_{t-2:t}\in\mathbb{R}^{258}$ and $d_z=8$, the oracle policy is
\begin{equation}
    \begin{gathered}
    \mathbf z_t^*=\tanh f_\varphi^{\mathrm{priv}}(\mathbf e_t)
        \in Z=[-1,1]^{d_z}\\
    \pi^{\mathrm{or}}_{\theta,\varphi}(\cdot\mid\mathbf x_t,\mathbf e_t)
        =\pi_\theta(\cdot\mid\mathbf x_t,\mathbf z_t^*)
    \end{gathered}
\label{eq:oracle}
\end{equation}
Both modules are jointly optimized for
\begin{equation}
    \max_{\theta,\varphi}\;J\!\left(\pi^{\mathrm{or}}_{\theta,\varphi}\right)
    \label{eq:oracle_objective}
\end{equation}
where $J$ is defined in Eq.~\eqref{eq:pomdp_objective}. PPO~\cite{schulman2017proximal} with an asymmetric critic optimizes both modules under the task objective, making the latent task-relevant.  

\textbf{Adaptation encoder learning.} With $f_\varphi^{\mathrm{priv}}$ and $\pi_\theta$ frozen, $f_\psi^{\mathrm{adapt}}$ estimates the interaction latent from $\mathbf h_t=\mathbf o_{t-99:t}\in O^{100}$, approximately a $2\,$s window at $50\,$Hz:
\begin{equation}
    \begin{aligned}
    \hat{\mathbf z}_t&=\tanh f_\psi^{\mathrm{adapt}}(\mathbf h_t)\in Z\\
    \mathbf a_t&\sim\pi_\theta(\cdot\mid\mathbf x_t,\hat{\mathbf z}_t)
    \end{aligned}
\label{eq:tma_policy}
\end{equation}
Rollouts generated using $\hat{\mathbf z}_t$ form the training
dataset $\mathcal D$, while the frozen privileged encoder
provides target latents $\mathbf z_t^*$. Only $\psi$ is updated by minimizing
\begin{equation}
    \mathcal L_{\mathrm{adapt}}(\psi)
    =\mathbb E_{(\mathbf h_t,\mathbf e_t)\sim\mathcal D}
    \left[\frac{1}{d_z}\left\|\hat{\mathbf z}_t-\mathbf z_t^*\right\|_2^2\right]
\label{eq:adapt_loss}
\end{equation}
Histories are padded with the first observation at episode start. As shown in Fig.~\ref{fig:method}(d), the privileged encoder is removed at deployment while the adaptation encoder updates $\hat{\mathbf z}_t$ from observed interaction history.

\subsection{Contact-Centric Domain Randomization}
\label{sec:randomization}

\textbf{Tactile randomization.} Sensor-frame calibration errors are modeled by perturbing $R_{\{WT_i\},t}$ during force transformation with orientation offsets fixed relative to the fingertip links. The filter coefficient $\beta$ is randomized to vary the temporal response. Following the cone-coordinate representation in Sec.~\ref{sec:tactile}, measurement noise is applied to the normal load $n_{i,t}$, force-cone angle $\alpha_{i,t}$, and shear direction $\phi_{i,t}$. Dropout and sensor failure are simulated by setting individual fingertip readings $\boldsymbol\tau_{i,t}$ to zero. These perturbations preserve the cone-coordinate structure while exposing the policy and adaptation encoder to tactile sensing variability.

\begin{figure*}[!t]
    \centering
    \includegraphics[width=0.95\textwidth]{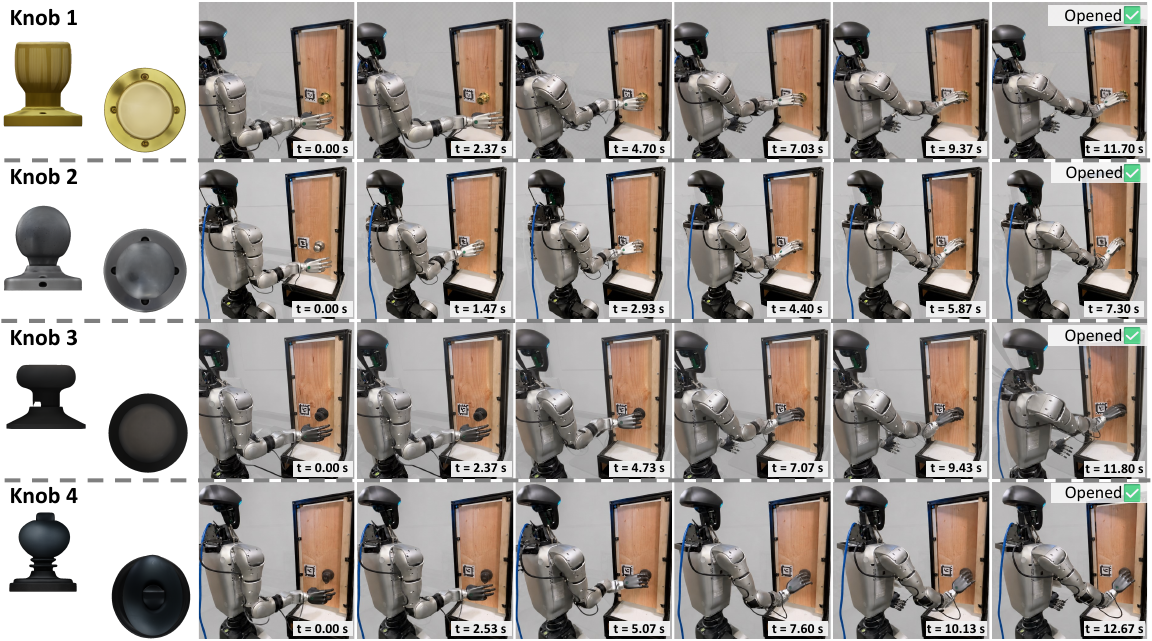}
\vspace{-0.5em}

    \caption{\textbf{Qualitative real-world rollouts of TMA.} Each row presents a commercial knob in front and top views, followed by a successful rollout from left to right: reaching, multi-finger grasping, knob rotation, and door opening. An AprilTag provides knob pose estimation, and tape applied to the knob rim increases contact friction. Timestamps indicate elapsed rollout time.}
    \label{fig:qualitative_results}
\vspace{-1em}
\end{figure*}

\textbf{Physical randomization.} Contact friction, spring stiffness $k_K$, preload $\tau_K^0$, knob and panel geometry, panel mass, and hand-actuation properties are randomized, together with the robot's initial configuration and pose. Selected hidden properties are provided to the privileged encoder through $\mathbf e_t^{\mathrm{dyn}}$ and $\mathbf e_t^{\mathrm{geo}}$. These variations diversify grasp geometry, required torque, door inertia, compliance, and reach conditions, encouraging adaptation from the combined interaction history.

%% file: text/04-experiments.tex
\section{EXPERIMENTAL RESULTS}

\subsection{Simulation Setup and Training}
\label{sec:simulation_setup}
We train with PPO in $8192$ parallel environments, with physics at $200\,$Hz and policy control at $50\,$Hz. Episodes last $20\,$s, including a $3\,$s settling period with manipulation increments suppressed. Privileged policy training takes approximately $15$ hours for $12{,}000$ iterations on a single NVIDIA RTX PRO 6000 GPU.

The simulated door includes a rotary knob, translating latch, and hinged panel. We randomize knob radius ($0.020$--$0.035\,$m), axial shape ratio ($0.4$--$1.0$), mounting height ($10\,$cm range), spring stiffness ($0.1$--$1.5\,$N\,m/rad), and preload ($0.01$--$0.30\,$N\,m). Contact friction ($0.4$--$1.5$), knob-to-hinge distance ($0.15$--$0.20\,$m), panel mass ($1$--$5\,$kg), hand actuation, and robot initialization are also randomized, together with the sensing perturbations in Sec.~\ref{sec:randomization}.

\begin{figure}[H]
    \centering
    \includegraphics[width=\columnwidth]{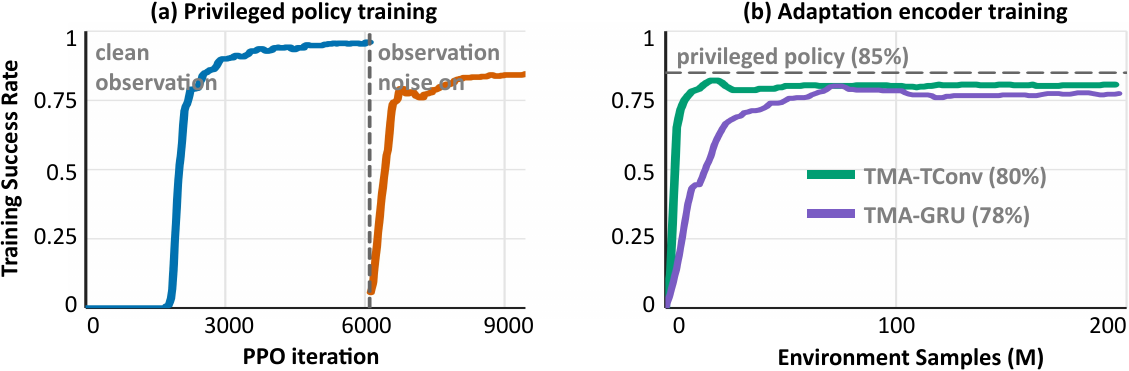}
    \caption{\textbf{Simulation training performance}. (a) The privileged policy is pretrained with clean observations and then fine-tuned with observation noise. (b) With the privileged policy frozen, TMA-TConv and TMA-GRU regress its interaction latent over the same $200$-million-sample window.}
    \label{fig:simulation_training}
\vspace{-1em}

\end{figure}

Following Sec.~\ref{sec:tma}, we jointly train the policy and privileged encoder with clean observations, then fine-tune with observation noise. Both modules are subsequently frozen while the adaptation encoder is trained through latent regression for $200$ million environment samples. We compare GRU~\cite{cho2014learning} and temporal-convolution (TConv)~\cite{bai2018empirical} encoders. TConv reaches strong task performance earlier and achieves slightly higher final success under the same sample budget (Fig.~\ref{fig:simulation_training}). We therefore use it in subsequent evaluations.

\subsection{Simulation Evaluation}
\label{sec:simulation_evaluation}

\textbf{Metrics.} For both simulation and real-world evaluation, we use two nested metrics to separate contact acquisition from complete task execution. A trial achieves \textit{grasp success} ($G$) when the thumb and at least two opposing fingers maintain tactile-observable knob contact for $1\,$s without a safety termination. Door-open success ($S$)  requires the hinge angle to exceed $15^\circ$ while maintaining the grasp ($S\leq G$).

\textbf{Signal ablation.} We evaluate in randomized environments without privileged inputs. TMA receives the complete cone-coordinate signal $\boldsymbol\tau_t$. Three variants replace it with the binary fingertip-contact mask $\mathbf c_t=[c_{i,t}]_{i=1}^{n_f}\in\{0,1\}^{n_f}$ formed from the indicators in Eq.~\eqref{eq:grip_gate}, the normal-load vector $\mathbf n_t=[n_{i,t}]_{i=1}^{n_f}$ from Eq.~\eqref{eq:tactile_encoding}, or the two tangential components $(f_{i,t}^x,f_{i,t}^y)$ of each calibrated tactile reading in Eq.~\eqref{eq:raw_tactile}, stacked across fingertips. All other settings are held fixed across three seeds, each with $4096$ evaluation episodes. 

\begin{table}[h]
    \vspace{-1em}
    \centering
    \caption{Simulation tactile-signal ablation. Input dimension denotes the per-frame tactile feature size. $G$ and $S$ denote grasp and door-open success rates. }
    \label{tab:simulation_comparison}
    \vspace{-1em}
    \footnotesize
    \setlength{\tabcolsep}{3pt}
    \begin{tabular}{lcccc}
        \hline
        Method & \shortstack{TMA \\ (Ours)} & \shortstack{Binary\\contact} & \shortstack{Normal\\load} & \shortstack{Shear\\components} \\
        \hline
        Input dims. & 16 & 4 & 4 & 8 \\
        $G$ (\%) & \textbf{91.3 \textpm{} 1.0} & $88.4 \pm 1.2$ & $86.5 \pm 1.1$ & $72.8 \pm 1.6$ \\
        $S$ (\%) & \textbf{80.9 \textpm{} 1.4} & $73.6 \pm 1.4$ & $74.2 \pm 1.5$ & $65.1 \pm 1.7$ \\
        \hline
    \end{tabular}
\vspace{-1em}
\end{table}

\textbf{Results.} Table~\ref{tab:simulation_comparison} shows that the complete tactile representation achieves the best performance in both grasp acquisition and door opening. Binary contact feedback supports grasp acquisition, but its lower opening success suggests that contact presence alone provides limited information for regulating the subsequent forceful interaction. Normal-load feedback also supports grasping, yet does not capture the tangential loading associated with knob rotation, which explains why it offers little improvement over binary contact in task completion. Shear-only feedback performs worst on both metrics, suggesting that tangential forces are difficult to interpret and regulate without the normal loading that supports frictional engagement. Together, these results indicate that normal and shear information play complementary roles: their combination supports both establishing a grasp and maintaining effective contact while actuating the mechanism.

\subsection{Real-World Evaluation}
\label{sec:real_world_evaluation}

\textbf{Baselines.} We compare TMA with two complementary baselines. RMA uses the same two-stage adaptation pipeline without tactile feedback. The reactive policy receives binary fingertip contact feedback over three observation frames and predicts actions without latent adaptation. These comparisons evaluate the complete TMA system against proprioception-based adaptation and short history binary tactile control.

\textbf{Test doorknobs.} We evaluate four commercial rotary knobs with distinct grasp geometries, shown in Fig.~\ref{fig:qualitative_results}: Knob 1 has a tapered barrel with decreasing diameter; Knob 2 is approximately spherical; Knob 3 is an axisymmetric disk with a rounded circumference; and Knob 4 has an elongated, olive-shaped profile. These geometries vary hand aperture, fingertip approach, contact distribution, and rotational leverage, while their distinct internal mechanisms add interaction-dynamics variation.

\textbf{Protocol.} Each method runs $10$ trials per knob with no real-world fine-tuning. Before each trial, the robot is manually placed in front of the door, varying its planar base pose, and the table height is varied over a $3.8\,$cm range. The protocol therefore also tests robot--object initialization variation.

\begin{figure}[h]
    \centering
    \includegraphics[width=0.92\columnwidth]{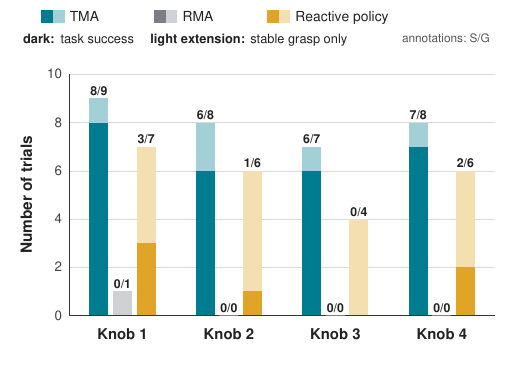}
\vspace{-2em}
    \caption{\textbf{Quantitative real-world results.} Each method is evaluated over \(10\) trials per knob.  Dark bar segments show door-open success $S$, and light segments show additional grasp-only successes. The total height is the number of grasp successes $G$. Annotations report $S/G$.}
    \label{fig:real_world_results}
\end{figure}

\begin{figure}[h]
    \centering
    \includegraphics[width=0.92\columnwidth]{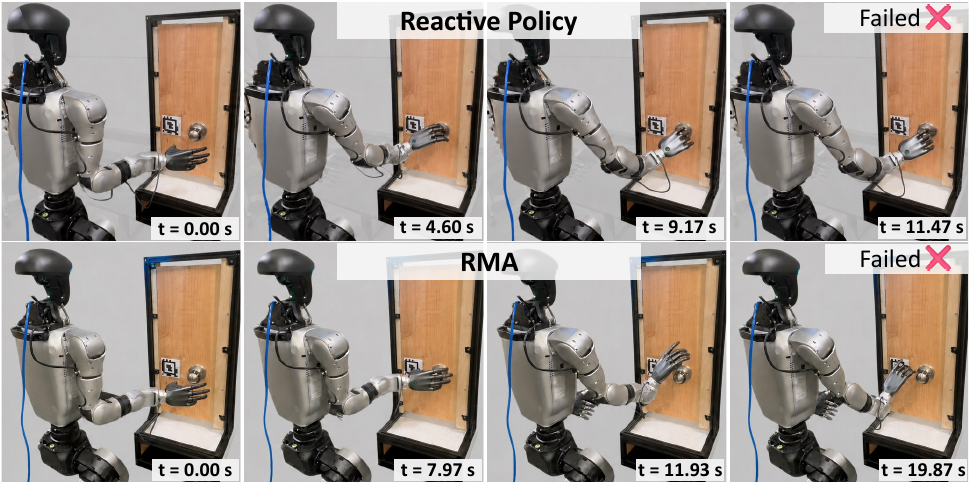}
    \caption{\textbf{Representative real-world baseline failures.} Top: the reactive policy, using binary-contact feedback over three observation frames, contacts the knob but fails to open the door. Bottom: RMA, using proprioceptive history without tactile feedback, fails to maintain engagement with the knob. Each sequence progresses from left to right with timestamps.}
    \label{fig:baseline_failures}
\vspace{-1em}
\end{figure}

\textbf{Results.} Fig.~\ref{fig:real_world_results} shows that TMA performs best across all four knobs, with most successful grasps progressing to door opening. This suggests that its advantage extends beyond contact acquisition to maintaining effective interaction during knob rotation and panel motion. The reactive policy frequently establishes a grasp but struggles to complete the task, indicating that contact presence alone provides insufficient feedback for the subsequent forceful interaction. Without force magnitude and direction, the policy cannot directly distinguish light contact from loading that sustains frictional engagement. RMA struggles even to maintain a grasp, with the hand drifting away from the knob after contact, as illustrated in Fig.~\ref{fig:baseline_failures}. This behavior is consistent with the limited contact information available through proprioception on a non-backdrivable hand, where changes in contact force may produce little joint displacement. Together, these comparisons support the combined use of force-aware tactile feedback and history-conditioned adaptation to sustain manipulation across variations in knob geometry, mechanism dynamics, and robot initialization.

%% file: text/05-results.tex
\section{DISCUSSION \& CONCLUSION}

Our results highlight three lessons for interaction-dominant humanoid manipulation. First, tactile feedback was essential for regulating local contact forces not directly observable from vision or proprioception, as RMA's poor performance showed that proprioceptive history was insufficient on a non-backdrivable hand. Second, instantaneous tactile measurements can be ambiguous, requiring interaction history to distinguish friction, grasp configuration, and mechanism dynamics through their responses to action. Third, the fingertip serves as both sensor and force-transmission interface. Since Coulomb friction bounds shear $s$ by $s \leq \mu n$, limited normal force $n$ and low friction $\mu$ can make the required knob torque physically unattainable. This explains the marked performance degradation after fingertip-pad damage. These findings were obtained through research experiments and do not represent a live production deployment. Future work will evaluate TMA on other interaction-dominant tasks and further ablate tactile feedback, temporal context, and the interaction objective.

\section*{ACKNOWLEDGMENT}
We thank Yuyin Sun for early discussions about the project, Hadi Tabatabaee for providing hardware support for the experiments, and Baljeet Singh for helpful discussions during our meetings.